\documentclass[11pt]{article}

\usepackage[
    paperwidth=8.5in,
    paperheight=11in,
    left=1in,
    right=1in,
    top=1in,
    bottom=1in
]{geometry}

\usepackage{modernpaper}
\usepackage{caption}
\usepackage{xurl}
\usepackage[numbers]{natbib}
\usepackage{algorithm}      % the floating "Algorithm" environment + caption
\usepackage{algpseudocode}  % \State, \For, \If, \Require, \Return, \Comment, etc.
\algrenewcommand\alglinenumber[1]{\footnotesize#1\hspace{0.8em}}

\usepackage{tikz}
\usetikzlibrary{tikzmark,calc,decorations.pathreplacing}
\usepackage{tikz}
\usetikzlibrary{arrows.meta,positioning,calc,fit,matrix}
\usepackage{pgfplots}
\pgfplotsset{compat=1.18}
\usetikzlibrary{shapes.geometric,shadows.blur}
\definecolor{nodefill}{HTML}{EFE7DC}\definecolor{nodeedge}{HTML}{444444}
\definecolor{ensfill}{HTML}{DCE6F2}\definecolor{ensedge}{HTML}{2C4B8E}
\definecolor{predfill}{HTML}{F6DCDC}\definecolor{prededge}{HTML}{C0392B}
\definecolor{datafill}{HTML}{ECECEC}\definecolor{dataedge}{HTML}{707070}
\definecolor{arrowcol}{HTML}{555555}\definecolor{wcol}{HTML}{C0392B}
\definecolor{distill}{HTML}{C0392B}
\newsavebox\panelLR
\newsavebox\panelChain
\newsavebox\panelSel
\usepackage{minted}
\usepackage{float}
\usepackage{graphicx}
\setminted{
    frame=single,
    framesep=8pt,
    baselinestretch=1.2,
    fontsize=\small,
    xleftmargin=12pt,
    breaklines,
}

\newcommand{\figorpending}[2][0.85\textwidth]{%
    \IfFileExists{#2}{\includegraphics[width=#1]{#2}}{%
        \fbox{\parbox{#1}{\centering\textit{[figure pending: \texttt{#2}]}\vspace{0.4em}}}%
    }%
}

\begin{document}

%% ============================================================================
%% TITLE
%% ============================================================================
\begin{center}
{\LARGE\bfseries q0: Primitives for Hyper-Epoch Pretraining}
\vspace{0.5cm}

{\large
Bishwas Mandal\textsuperscript{1}\quad
Shmuel Berman\textsuperscript{2}\quad
Akshay Vegesna\textsuperscript{1}\quad
Samip Dahal\textsuperscript{1}
}

\vspace{0.25cm}
{\small\textsuperscript{1}Q Labs \quad \textsuperscript{2}Princeton University}

\vspace{0.1cm}
\end{center}

\renewcommand{\thefootnote}{\fnsymbol{footnote}}
\footnotetext[1]{See author contributions at the end of the paper. Correspondence: \texttt{research@qlabs.sh}}
\renewcommand{\thefootnote}{\arabic{footnote}}

\begin{abstract}
% Multi-epoch training is becoming the standard now that compute is growing
% faster than the supply of high-quality text. Repeating epochs over the same
% data stops helping after a few passes, so we ask how the remaining budget is
% best used. We argue this budget is better used exploring many models and
% weighting them than refining a single model, and introduce hyper-epoch
% pretraining (\emph{q0}), which turns a multi-epoch
%  budget into a diverse population of models whose combined
%  predictions reach a lower validation loss than a single refined model. A cyclic schedule with anti-correlated learning rate and weight decay
% collects diverse snapshots from a few parallel trajectories. Chain distillation
% trains each snapshot against its predecessor so that member quality compounds.
% A learned prior, fit on a held out set, selects and weights members for any
% inference budget. On a 1.8B-parameter model trained on 100M FineWeb tokens,
% \emph{q0} matches a strong 256-epoch ensemble baseline using only ${\sim}56$
% epochs (${\sim}4.6\times$ fewer), or ${\sim}67$ epochs (${\sim}3.8\times$ fewer)
% even when matched to the baseline's ensemble size, and continues to
% improve beyond it. These gains transfer to downstream accuracy on ARC-Easy,
% PIQA, and SciQ, and reach cumulative ${\sim}12.9\times$ data efficiency under
% the Slowrun setting. Crucially, the optimal allocation shifts with the budget,
% so we give prescriptive recipes for how to spend a given epoch budget to
% maximize generalization, from a \emph{single
% epoch} up to the largest budgets.

Multi-epoch training is becoming the standard now that compute is growing
faster than the supply of high-quality text. But pretraining a single model saturates
within a few passes, long before the compute budget is exhausted. We argue
this calls for a \emph{conceptual shift} from training a
single model toward exploring a population of models and aggregating their
predictions. We introduce hyper-epoch pretraining (\emph{q0}), which turns a
multi-epoch budget into a population of diverse models whose combined
predictions reach a lower validation loss than a single refined model. \emph{q0} reduces to three core primitives. A cyclic
schedule with anti-correlated learning rate and weight decay collects diverse
models from a few parallel trajectories. Chain distillation trains each model
against its predecessor so that model quality compounds across the population.
A learned prior, fit on a held out set, selects and weights members for any
inference budget. On a 1.8B-parameter model trained on 100M FineWeb tokens,
\emph{q0} matches a strong 256-epoch ensemble baseline using only ${\sim}56$
epochs (${\sim}4.6\times$ fewer), or ${\sim}67$ epochs (${\sim}3.8\times$ fewer)
when matched to the baseline's ensemble size, and continues to improve
beyond it. These gains reach cumulative ${\sim}12.9\times$ data efficiency under
the Slowrun setting and transfer to downstream benchmarks. Crucially, the
optimal allocation shifts with the budget, so we give prescriptive recipes for
how to spend a given epoch budget to maximize generalization, from a
single epoch up to the largest budgets.

\vspace{1.5em}

{\centering
\includegraphics[width=1.0\linewidth]{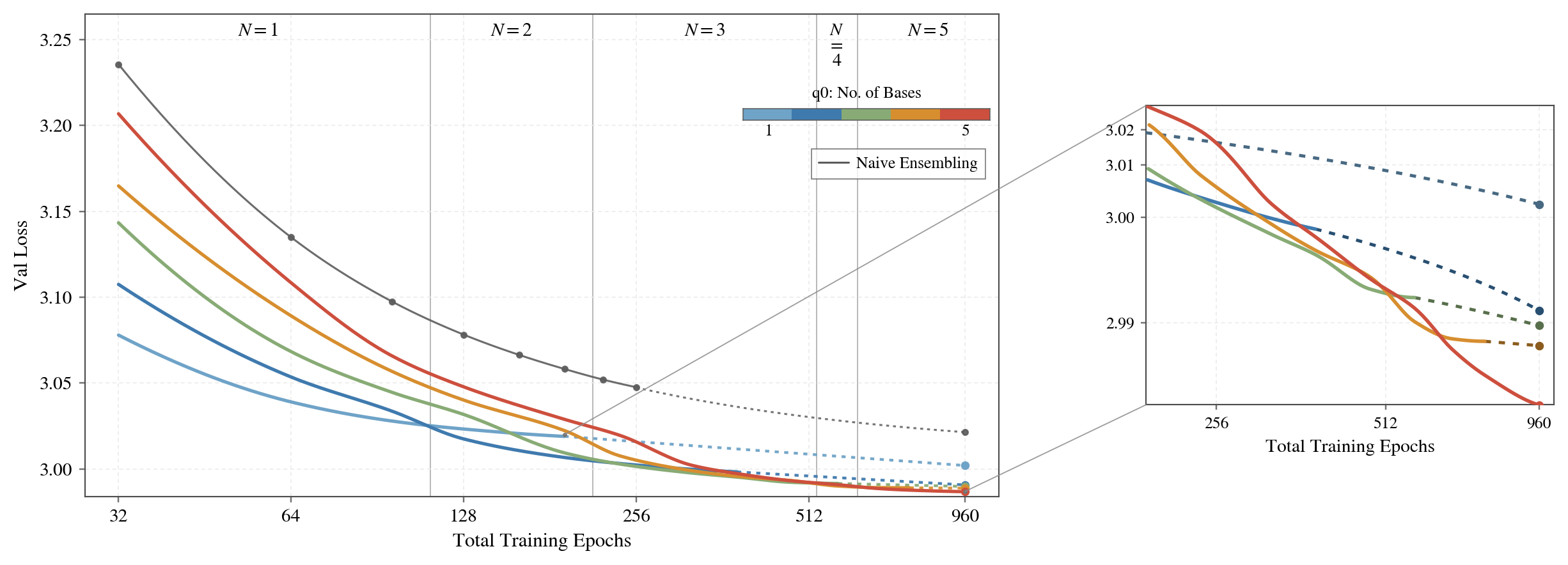}
\captionof{figure}{\emph{q0 converges substantially faster and to a lower loss across all epoch budgets, and the advantage is largest in the practically relevant small-to-medium-epoch regime where most training actually operates.}}
\label{fig:main}
\par}
\end{abstract}

\vspace{0.11 cm}

\section{Introduction}
\label{introduction}

Progress in language modeling has largely come from scaling compute and data together \citep{kaplan2020scaling, hoffmann2022training}. The supply of high quality text, however, is fundamentally limited while compute continues to grow, and frontier models have already consumed a substantial fraction of available data \citep{muennighoff2023scaling, villalobos2024position}. As a result, scaling is increasingly entering a data-constrained regime, where further progress depends on how additional computation is used on a fixed corpus. Multi-epoch training is a natural usage of increased compute. However, performing repeated passes over the same data exhibits diminishing returns as models converge and doesn't improve model capability after a few epochs \citep{lovelace2026prescriptivescalinglawsdata, kim2025infinite}. This raises the question of how compute should be allocated given a fixed dataset and a budget of $N$ training epochs.

We approach this question from first principles. Solomonoff induction, a foundational framework for generalization, suggests that one should consider a large space of hypotheses explaining the observed data and weight them according to a complexity prior \citep{solomonoff1964formal, hutter2005universal}. It views prediction as averaging over all computable explanations of the data, weighted by their description length or simplicity. This perspective naturally benefits from increased compute, as additional resources enable broader exploration over the hypothesis space. The natural implication is a shift in how a hyper-epoch budget is used: instead of refining a single model long past the point where extra epochs help, we explore a diverse population of models on the same data and aggregate their predictions
at inference.

% \citep{garipov2018loss, draxler2018essentially}

% Ensembling \citep{lakshminarayanan2017simple, fort2019deep, kim2025infinite} is the most direct realization of this idea, but a crude one. First, at any realistic compute budget, we can train only a limited number of models, and most compute is wasted training full models from scratch. Second, independently trained models tend to be similar in quality and adding more of them does not compound individual model capability. Third, the resulting models are combined with uniform weights, ignoring that some models might be more useful than the others. Our setting, \emph{hyper-epoch pretraining}, also referred to as \emph{q0}, addresses each of these in turn.

Ensembling \citep{lakshminarayanan2017simple, fort2019deep, kim2025infinite} is
the most direct realization of this idea, but a crude one. We argue that building
a population of models requires three things, and naive ensembling falls short on
each. First, we need a cheap way to explore many models, yet ensembling trains
every model from scratch, so most compute is wasted and only a handful of models
are affordable. Second, the models should compound in capability, yet
independently trained models are similar in quality, so adding more does not lift
individual capability. Third, their predictions should be weighted to maximize
generalization, yet ensembling weights every model uniformly, ignoring that some
are more useful than others. Everything in \emph{q0} follows from this one goal of
efficiently training a population of models, and we contribute a primitive for each
requirement.

\textbf{Faster exploration of weight space.}
To explore broadly without retraining every model from scratch, we adopt a strategy inspired by snapshot ensembles \citep{huang2017snapshot} and Fast Geometric Ensembling (FGE) \citep{garipov2018loss}. Our approach is closer to FGE, as we use short training cycles to collect many models
along a single cyclic training trajectory. We
further train a small number of such trajectories from different random
initializations to add diversity across them. As a result, only a few models
are trained fully from scratch.

\textbf{Capability compounding across the population.}
Standard gradient descent under identical data and compute budgets produces
models of similar quality to one another, regardless of how many we train
\citep{bouthillier2021accounting, summers2021nondeterminism}. We introduce
\emph{chain distillation}, where each new snapshot is trained against the
previous one as a teacher in addition to the standard objective. Each successive
model conditions on its predecessor and improves on it, so the population's
capability compounds across training rather than producing models of similar
individual quality.

\textbf{Learned weighting at inference.}
We replace uniform averaging with a weighting learned on a held-out fitness set.
While Solomonoff's prior weighs hypotheses by compressibility and Bayesian model
averaging by posterior plausibility \citep{mackay1992bayesian, neal1996bayesian},
both are intractable in modern settings, so we adopt this simpler empirical
proxy.

Together these three primitives constitute hyper-epoch pretraining: a framework for allocating compute over a fixed dataset by training for hundreds of epochs to produce a diverse, mutually informed population of models. On a 1.8B parameter model trained on 100M FineWeb tokens, \emph{q0} matches a strong 256 epoch ensemble baseline at  56 epochs, corresponding to a $4.6\times$ reduction in training epochs ($3.8\times$ reduction even at the baseline's matched ensemble size), and continues to improve beyond this point. We further observe that these improvements in pretraining transfer to downstream benchmarks such as ARC-Easy, PIQA, and SciQ. Under the Slowrun setting\footnote{\url{https://github.com/qlabs-eng/slowrun}},
\emph{q0} improves data efficiency to $12.9\times$ on validation
loss relative to the baseline, and to $16\times$ in average
downstream accuracy.

We find that these gains persist across all epoch budgets, from as low as 1 to as high as 960, and that optimal allocation varies systematically with scale, with different budgets favoring different distributions of training across trajectories and within-cycle dynamics. Notably, the learned generalization prior exhibits non-trivial behavior, assigning substantial weight to snapshots that are not individually optimal under validation loss, but are complementary for ensembling.

%% ============================================================================
\section{Methodology}
\label{sec:method}
We now formalize hyper-epoch pretraining and give the mechanism behind each of
the three primitives introduced earlier. We denote the population by
$\{\theta_{n,c}\}_{n=1,\,c=1}^{N,\,C}$, where $n \in \{1,\dots,N\}$ indexes a
parallel trajectory from an independent random initialization and
$c \in \{1,\dots,C\}$ indexes a cycle within that trajectory, for a total of
$M = N \cdot C$ models. We develop each primitive in turn: the cyclic,
parallelized schedule that yields the snapshots $\theta_{n,c}$
(§\ref{sec:method-snapshots}); the chain-distillation term that couples each
snapshot to its predecessor (§\ref{sec:method-chain-distill}); and the learned
prior that selects and weights the population at inference
(§\ref{sec:method-weighting}). An overview of our methodology is shown in Figure \ref{fig:schedule}.

\subsection{Fast Exploration of Weight Space via Snapshot Ensembling}
\label{sec:method-snapshots}
Our first primitive collects a large, diverse population of models without
training $M$ of them from scratch. Instead, we run a small number of
parallel trajectories and save several snapshots trained with cyclic schedules that favor exploration along each one, so the cost
of a single full run is shared across many ensemble members.

Within each trajectory, training is split into $C$ cycles. Each cycle
restarts the learning rate at a high value and anneals it down to a small
floor, and at the end of the cycle we save the parameters as a snapshot
$\theta_{n,c}$.
Intuitively, annealing the learning rate to a small value is the standard way to reach a
good final loss, so ending each cycle low ensures every snapshot has settled
into a local minimum rather than stopping at an arbitrary point partway
down. Restarting the learning rate high at the start of the next cycle then
kicks the model out of that minimum and pushes it to explore a different
region of the loss landscape. Repeating this rise and fall is what produces
diversity: the snapshots are individually well converged, yet different
enough from one another to be worth ensembling.

We further anti-correlate weight decay with the learning rate
(Eq.~\ref{eq:cyclic-schedule}). The high-LR, low-WD phase favors exploration,
while the low-LR, high-WD phase is intended to draw the parameters into a
low-norm basin just before the snapshot is taken. This concentrates each
cycle into two distinct regimes, explore and then settle, so that the
snapshot reflects the basin the model exploited rather than a point along the
way.

Figure~\ref{fig:schedule}(a) contrasts our schedule with a standard
single-cycle cosine reference. At the same total epoch budget, the standard
schedule yields one final model after the equivalent of $C$ cycles, whereas
ours yields $C$ snapshots, one per cycle.

\subsection{Model Capability Compounding via Chain Distillation}
\label{sec:method-chain-distill}

Snapshots from \S\ref{sec:method-snapshots} are still supervised only
by one-hot labels, so per-snapshot quality is bounded by what
cross-entropy can extract from those labels alone. Since ensemble
error depends on both member error and member disagreement, lifting
per snapshot quality directly strengthens the ensemble. Our second
primitive routes a richer signal to each new snapshot by treating the
preceding snapshot in the same trajectory as a frozen teacher: after
a warmup of $c_{\mathrm{start}}$ cycles, every subsequent cycle
$c > c_{\mathrm{start}}$ trains $\theta_{n,c}$ with
\begin{equation}
\label{eq:chain-distill}
    \mathcal{L}_c
    \;=\;
    (1 - \alpha)\,\mathcal{L}_{\mathrm{CE}}\!\bigl(\theta_{n,c}\bigr)
    \;+\;
    \alpha \, T^2 \cdot
    \mathrm{KL}\!\left(
        \sigma\!\bigl(f(x;\theta_{n,c-1}) / T\bigr)
        \,\Big\|\,
        \sigma\!\bigl(f(x;\theta_{n,c}) / T\bigr)
    \right),
\end{equation}
where $f(\cdot;\theta)$ are the pre-softmax logits, $\sigma$ the
softmax, $T$ the temperature, and $\alpha$ trades off labels against
teacher targets. The teacher $\theta_{n,c-1}$ is frozen during cycle
$c$ and refreshed at the end of every cycle. Refer to Figure \ref{fig:schedule} (b).

Why does the predecessor's distribution help when the predecessor is
itself trained from the same data? On the face of it no new
information has been introduced and theoretical work on training
LMs on their own outputs predicts that recursive self-training should
in fact degrade the model
\citep{shumailov2024curse}. Empirically the opposite is observed
across a wide range of setups: born again networks (BANs)
\citep{furlanello2018born} train a sequence of fresh models with each
successor receiving its predecessor's full output distribution and
routinely outperform their teacher; self-distillation more broadly
shows the same effect \citep{mobahi2020selfdistillation}; and recent
work on language models finds that augmenting training with the
model's own generations \citep{zelikman2022star,gulcehre2023rest} or
with synthetic data produced by a same class teacher
\citep{gunasekar2023textbooks} consistently improves the student
rather than collapsing it. The gradient analysis of
\citep{furlanello2018born} pinpoints two complementary mechanisms in
the soft target case: a \emph{dark knowledge} signal in the teacher's
wrong-class probabilities, which exposes inter-class similarity
structure that one-hot labels cannot transmit, and an implicit
\emph{importance weighting} in which the teacher's confidence on the
correct class rescales each example's gradient. Both make supervision
strictly richer than labels alone. Our setting is closest to BANs where we use
soft targets rather than sampled tokens and we apply the same
mechanism within the trajectory rather than across runs.

% Distillation also pulls consecutive snapshots together, so
% within trajectory disagreement drops as we observe this empirically.
% The ensemble nevertheless improves because the gain in individual
% quality outweighs the loss in diversity, as the error–ambiguity decomposition \citep{krogh1995neural} predicts. 

\begin{algorithm}[!t]
\caption{q0: Hyper-epoch pretraining}
\label{alg:full}
\begin{algorithmic}[1]
\Require trajectories $N$, cycles per trajectory $C$, warmup
    $c_{\mathrm{start}}$, distill weight $\alpha$, temperature $T$,
    cyclic LR/WD schedule; fitness set $\mathcal{F}$; optimization
    steps $T_{\mathrm{opt}}$, learning rate $\eta_\beta$, weight
    decay; inference budget $K \le M$
\For{$n = 1, \dots, N$ \textbf{in parallel}}
    \State initialize
        $\theta \leftarrow \theta_{n,0}$ randomly;\quad
        $\theta_{\mathrm{teacher}} \leftarrow \texttt{None}$
    \For{$c = 1, \dots, C$}
        \For{each step in cycle $c$}
            \State apply cyclic LR/WD multipliers
                (Eq.~\ref{eq:cyclic-schedule})
            \If{$c \leq c_{\mathrm{start}}$ \textbf{or}
                $\theta_{\mathrm{teacher}}$ is $\texttt{None}$}
                \State $\mathcal{L} \leftarrow
                    \mathcal{L}_{\mathrm{CE}}(\theta)$
            \Else
                \State $\mathcal{L} \leftarrow
                    (1{-}\alpha)\,\mathcal{L}_{\mathrm{CE}}(\theta)
                    + \alpha T^2\,
                    \mathrm{KL}\!\bigl(\sigma(f(x;\theta_{\mathrm{teacher}})/T)\,\|\,
                    \sigma(f(x;\theta)/T)\bigr)$
            \EndIf
            \State update $\theta$ with optimizer step on $\mathcal{L}$
        \EndFor
        \State $\theta_{n,c} \leftarrow \theta$;\quad
            $i \leftarrow (n-1)\,C + c$
            \Comment{snapshot}
        \For{each token $t \in \mathcal{F}$}
            \State $P_{\mathcal{F}}[i,\,t] \leftarrow
                \sigma\!\bigl(f(x_{<t};\,\theta_{n,c})\bigr)_{y_t}$
                \Comment{cache per-token $p(\text{gt})$}
        \EndFor
        \State $\theta_{\mathrm{teacher}} \leftarrow \theta_{n,c}$
            \Comment{refresh teacher}
    \EndFor
\EndFor
\State $M \leftarrow N \cdot C$;\quad
    $\{\theta_i\}_{i=1}^{M} \leftarrow \{\theta_{n,c}\}_{n,c}$
\State initialize $\boldsymbol{\beta} \leftarrow
    10^{-4}\,\mathcal{N}(0, I_M)$
\For{$\tau = 1, \dots, T_{\mathrm{opt}}$}
    \State $w \leftarrow \mathrm{softmax}(\boldsymbol{\beta})$;\quad
        $q \leftarrow w^{\top} P_{\mathcal{F}}$
    \State $\mathcal{L}(\boldsymbol{\beta}) \leftarrow
        -\frac{1}{|\mathcal{F}|}\sum_t \log q_t$
        \Comment{Eq.~\ref{eq:learned-prior}}
    \State $\boldsymbol{\beta} \leftarrow$ AdamW step on
        $\nabla_{\boldsymbol{\beta}} \mathcal{L}$
\EndFor
\State $w^\star \leftarrow \mathrm{softmax}(\boldsymbol{\beta})$;\quad
    $\mathcal{S}_K \leftarrow$ top-$K$ indices of $w^\star$;\quad
    $\tilde w_i \leftarrow
    w^\star_i \,/\, \sum_{j \in \mathcal{S}_K} w^\star_j$
\State \Return $p_{\mathrm{ens}}(x) =
        \sum_{i \in \mathcal{S}_K} \tilde w_i \, p_i(x)$
\end{algorithmic}
\end{algorithm}

\subsection{Learned Generalization Prior}
\label{sec:method-weighting}

After training we have $M$ snapshots, but at inference each ensemble
member costs one forward pass. For any inference budget $K \le M$ we
therefore need to decide \emph{which} $K$ snapshots to keep and
\emph{how to weight} them in the combination. These choices are coupled: the optimal subset depends on the weighting, and the optimal weighting depends on the subset, so picking members and weights separately is  generically suboptimal. Our third
primitive solves both choices jointly, for every $K$ at once, with a
single gradient fit on a held out fitness set.

Let $\mathcal{F}$ be a small held out \emph{fitness} set of the
training distribution that no snapshot is ever trained on. We parameterize a
distribution over the $M$ snapshots as a softmax over unconstrained
logits $\boldsymbol{\beta} \in \mathbb{R}^M$ and minimize the ensemble's mean
negative log-likelihood on $\mathcal{F}$:
\begin{equation}
\label{eq:learned-prior}
    \mathcal{L}(\boldsymbol{\beta})
    \;=\;
    -\frac{1}{|\mathcal{F}|}
    \sum_{t \in \mathcal{F}}
    \log\!\left(
        \sum_{i=1}^{M} w_i \, p_{i,t}
    \right),
    \qquad
    w \;=\; \mathrm{softmax}(\boldsymbol{\beta}),
\end{equation}
where $p_{i,t}$ is the probability that snapshot $\theta_i$ assigns
to the ground truth token at position $t$. The $w_i$ appear
\emph{inside} the log and we mix probabilities, not log-probabilities so the loss is not separable over members.

% %: snapshots that are individually strong but redundant with the rest of the population receive small weight, and snapshots that are individually weaker but
% cover errors others miss receive large weight.%

The fitted $w$ then serves any inference budget by truncation: for
each target $K \le M$, we keep the $K$ snapshots with the largest
fitted weights and combine them with their weights renormalized to
sum to one. The same fit is reused across all values of $K$ i.e.,
we do not re-fit per budget. Conceptually, this is motivated by a Solomonoff style
complexity prior over hypotheses: we cannot compute Kolmogorov
complexity, but we can empirically compute held out generalization on
training distribution data and use that as the weighting signal. Algorithm~\ref{alg:full} states the complete procedure.

%% ============================================================================
\section{Experiment and Results}
\label{sec:experiments}

Our experiments use the model and setting from the Slowrun challenge: a
1.8B parameter decoder-only transformer trained on 100M tokens of FineWeb
\citep{penedo2024the} and evaluated on a held out 10M token validation set from
the same distribution. The model is heavily regularized and overparameterized, which is the configuration that performs best in
this data constrained regime and therefore makes for a strong baseline. In our primary configuration, we run $N=2$ parallel trajectories (base models) over $C=64$ cycles, where each cycle spans $2$ epochs, yielding $M=128$ snapshots in total, and $256$ total epochs. Details of the architecture, optimizer, and complete training hyperparameter sweep are provided in Appendix \ref{app:hyperparams}.

We report validation cross-entropy
loss on the held out FineWeb data, and zero-shot accuracy on ARC-Easy \citep{Clark2018ThinkYH}, PIQA \citep{Bisk2020}, and SciQ \citep{welbl-etal-2017-crowdsourcing}. Ensembles combine
member predictions in probability space using the learned
generalization prior of \S\ref{sec:method-weighting}.

\begin{figure}[!t]
    \centering
    \includegraphics[width=0.6\textwidth]{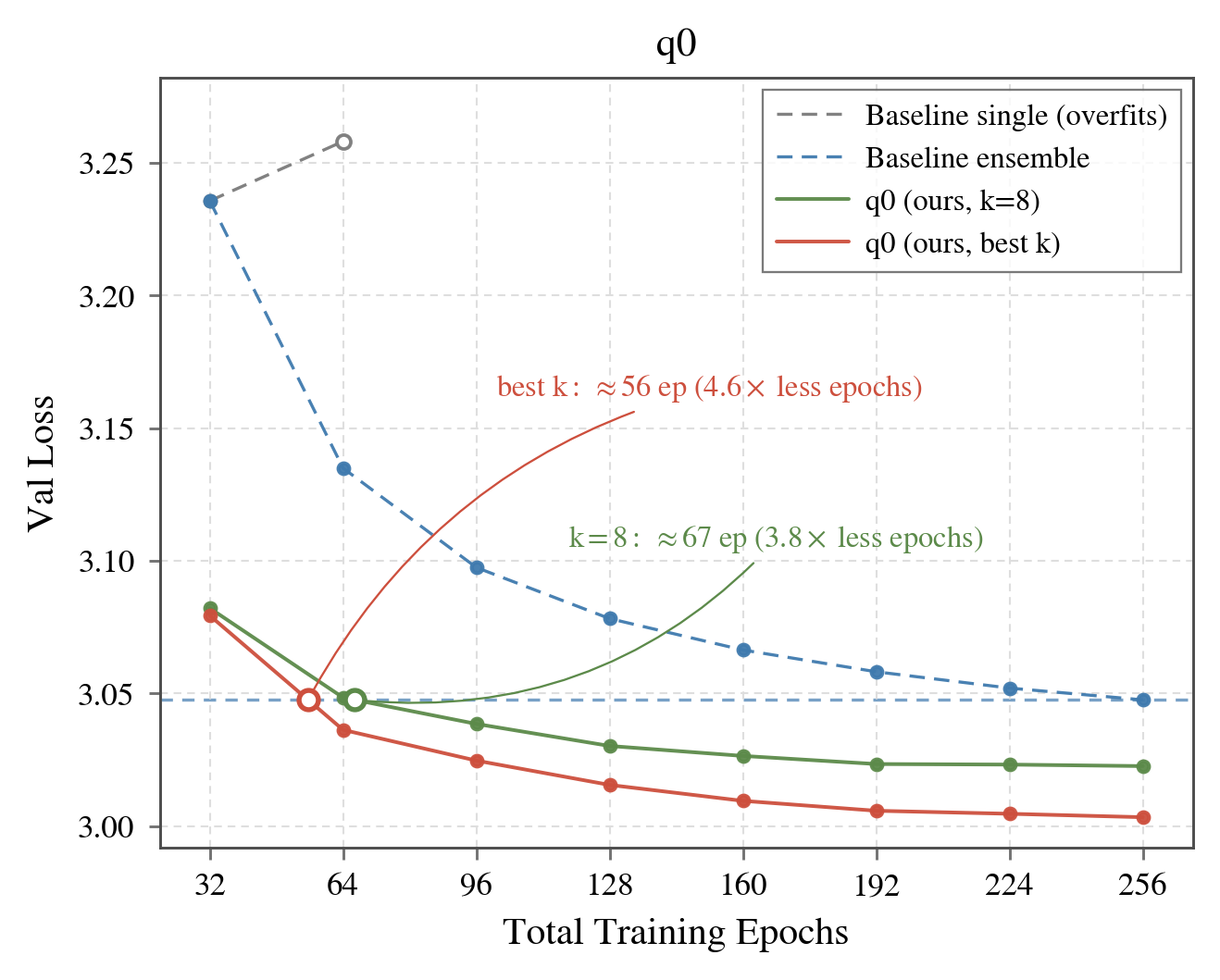}
    % \caption{Validation loss vs.\ total training epochs for the baseline
    % (independent runs with EMA and naive averaging) and q0:
    % hyper-epoch pretraining. Our technique matches the $256$ epoch baseline at
    % ${\sim}56$ epochs (${\sim}4.6\times$ efficiency) and
    % is consistently better throughout the $256$ epochs.}
    \caption{Validation loss vs.\ total training epochs for the baseline
      (independent runs with EMA and naive averaging) and q0:
      hyper-epoch pretraining, shown both with best-$k$ member selection
      and at a fixed $k{=}8$. Our best-$k$ technique matches the $256$ epoch
      baseline at ${\sim}56$ epochs (${\sim}4.6\times$ efficiency); even at a
      fixed $k{=}8$ it matches the baseline at ${\sim}67$ epochs
      (${\sim}3.8\times$ efficiency). Both are consistently better than the
      baseline throughout the $256$ epochs. Crossover epochs are read off by
      linear interpolation between the bracketing measured points.}
      \label{fig:main-result}
    % \label{fig:main-result}
\end{figure}

For baselines, we train $8$ independent models from random initialization for $32$ epochs each ($256$ epochs total), with EMA weight averaging and no cyclic LR/WD snapshot schedules or chain distillation, and combine them at inference by uniform softmax averaging. Additionally, we evaluate both the baseline and \emph{q0} under the data-efficiency setting from the Slowrun challenge; details are deferred to Appendix~\ref{app:slowrun}.

\subsection{Main Results: Hyper-Epoch Pretraining}
\label{sec:main-result}

In figure~\ref{fig:main-result}, we compare our hyper-epoch pretraining (q0)
against the strongest independent runs baseline across the $32$ to $256$ epoch
regime. We adopt $8$ models trained for $32$ epochs each as our baseline. At the same
total budget of $256$ epochs, the alternative split of $16$ models at $16$
epochs each gave worse ensemble results. q0 (best $k$) matches the $256$ epoch baseline at only ${\sim}56$ epochs of training, a ${\sim}4.6\times$ epoch reduction, and continues to improve
  thereafter, reaching a final validation loss of $3.003$ at $256$ epochs. For
  each compute budget $E$, q0 reports the better of two configurations: training
  one base model for $E$ epochs, or training two base models for $E/2$ epochs
  each. Both are feasible up to $E=128$; above this point only the 2 base
  configuration is feasible, since a single base model can be trained for at most
  $128$ epochs in our setup. Both q0 curves differ only in the ensemble size $k$. We report a fixed $k{=}8$ to match the baseline ensemble, which combines $8$ models; even at $k{=}8$, q0 matches the $256$ epoch
  baseline at ${\sim}67$ epochs (a ${\sim}3.8\times$ reduction). Table~\ref{tab:main-result} reports the numerical comparison at matched epoch budget and also shows the result of our scaled variant, which reaches a
validation loss of $2.987$ at $960$ total epochs. The final column
reports data efficiency under the Slowrun setting
(Appendix~\ref{app:slowrun}), our baseline already reaches
${\sim}10\times$ data efficiency at 256 epochs relative to the nanochat
baseline, q0 lifts this to ${\sim}12\times$ at matched budget, and
the scaled variant pushes it further to ${\sim}12.9\times$.

\begin{table}[!t]
\centering
\caption{Comparison of the strongest baseline against q0
  (hyper-epoch pretraining). For q0 we report the best top-$K$ snapshot
  ensemble obtained by gradient fit (Sec.~\ref{sec:method-weighting});
  the optimum is $K=128$, while $K\in\{16, 32, 64\}$ are only
  marginally weaker and still substantially outperform the baseline.
  Data efficiency is computed under the Slowrun challenge numbers
  (Appendix~\ref{app:slowrun}).}
\label{tab:main-result}
\vspace{0.5em}
\begin{tabular}{@{}lccc@{}}
\toprule
\textbf{Approach}  & \textbf{Total Epochs}  & \textbf{Val Loss} & \textbf{Data Eff.} \\
\midrule
Baseline          & $256$           & $3.0476$          & ${\sim}10\times$ \\
q0 (ours)        & $256$     & $3.0034$          & ${\sim}12\times$ \\
q0 (ours, scaled)   & $960$ & $\textbf{2.9870}$  & $\mathbf{{\sim}12.9\times}$ \\
\bottomrule
\end{tabular}
\end{table}
%($8$ models $\times$ $32$ epochs, EMA)%

\begin{table}[H]
  \centering
    \caption{Zero-shot downstream accuracy on the three benchmarks, with data efficiency in grey (Appendix~\ref{app:slowrun-downstream}). q0 rows use the same top-$K$ weighted snapshot ensemble as in Table~\ref{tab:main-result} (selected by gradient fit; Sec.~\ref{sec:method-weighting}), and q0
  outperforms the strongest baseline across all three. Standard errors are on the order of $\pm 0.010$--$0.013$ and are omitted for brevity.}
  \label{tab:benchmarks}
  \vspace{0.5em}
  \begin{tabular}{@{}lcccc@{}}
  \toprule
  \textbf{Approach} & \textbf{ARC-Easy} & \textbf{PIQA} & \textbf{SciQ} &
  \textbf{Average} \\
  \midrule
  Baseline
      & $0.4781$ {\scriptsize\textcolor{gray}{$7.9\times$}}
      & $0.6697$ {\scriptsize\textcolor{gray}{$8.0\times$}}
      & $0.7850$ {\scriptsize\textcolor{gray}{$13.4\times$}}
      & $0.6443$ {\scriptsize\textcolor{gray}{$12.2\times$}} \\
  q0 (ours)
      & $0.4865$ {\scriptsize\textcolor{gray}{$12.7\times$}}
      & $\mathbf{0.6823}$ {\scriptsize\textcolor{gray}{$17.0\times$}}
      & $0.8030$ {\scriptsize\textcolor{gray}{$18.8\times$}}
      & $0.6573$ {\scriptsize\textcolor{gray}{$14.2\times$}} \\
  q0 (ours, scaled)
      & $\mathbf{0.4945}$ {\scriptsize\textcolor{gray}{$13.2\times$}}
      & $0.6774$ {\scriptsize\textcolor{gray}{$14.7\times$}}
      & $\mathbf{0.8080}$ {\scriptsize\textcolor{gray}{$19.9\times$}}
      & $\mathbf{0.6600}$ {\scriptsize\textcolor{gray}{$16.0\times$}} \\
  \bottomrule
  \end{tabular}
  \end{table}

The pretraining validation loss improvements
transfer to downstream benchmarks. In Table~\ref{tab:benchmarks}, we report
zero-shot accuracy on three commonly used benchmarks (at this scale): ARC-Easy,
PIQA, and SciQ. q0 outperforms the strongest baseline on every task. Translating
these accuracies into data efficiency (Appendix~\ref{app:slowrun-downstream}),
the baseline reaches ${\sim}12.2\times$ on average accuracy at 256 epoch budget, q0 lifts this to
${\sim}14.2\times$ at the same budget, and the scaled variant reaches ${\sim}16.0\times$. Per-benchmark data efficiency numbers are shown in grey in the table.

% \begin{table}[H]
%   \centering
%   \caption{Zero-shot downstream accuracy ($\pm$ standard error) on
%   ARC-Easy, PIQA, and SciQ. q0 rows
%   use the same top-$K$ weighted snapshot ensemble as in
%   Table~\ref{tab:main-result} (selected by gradient fit;
%   Sec.~\ref{sec:method-weighting}). q0 outperforms the strongest baseline
%   across all three benchmarks. The final column reports data efficiency on
%   average accuracy under the Slowrun setting (Appendix~\ref{app:slowrun-downstream}).}
%   \label{tab:benchmarks}
%   \vspace{0.5em}
%   \begin{tabular}{@{}lccccc@{}}
%   \toprule
%   \textbf{Approach} & \textbf{ARC-Easy} & \textbf{PIQA} & \textbf{SciQ} &
%   \textbf{Average} & \textbf{Data Eff.} \\
%   \midrule
%   Baseline
%       & $0.4781\pm0.010$ & $0.6697\pm0.011$ & $0.7850\pm0.013$ & $0.6443$
%       & ${\sim}12.2\times$ \\
%   q0 (ours)
%       & $0.4865\pm0.010$ & $\mathbf{0.6823\pm0.011}$
%       & $0.8030\pm0.013$ & $0.6573$ & ${\sim}14.2\times$ \\
%   q0 (ours, scaled)
%       & $\mathbf{0.4945\pm0.010}$ & $0.6774\pm0.011$
%       & $\mathbf{0.8080\pm0.013}$ & $\mathbf{0.6600}$
%       & $\mathbf{{\sim}16.0\times}$ \\
%   \bottomrule
%   \end{tabular}
%   \end{table}

\subsection{Optimal Compute Allocation}
\label{sec:compute-allocation}

% The best choice of
% $\tau$ turns out to be regime-dependent: in the small-budget regime
% ($E\!\le\!4$) end-loaded schedules with short, irregular cycles win
% (see Table~\ref{tab:small-budget}), whereas 

\begin{figure}[!t]
    \centering
    \includegraphics[width=\textwidth]{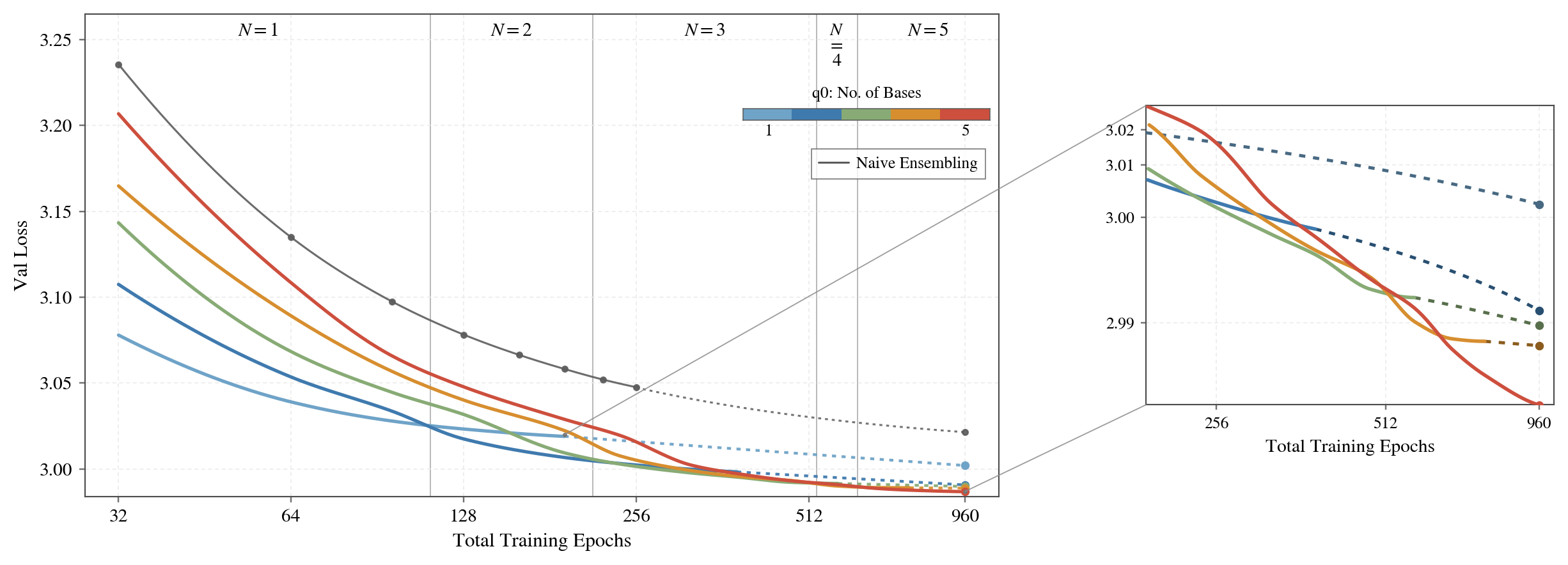}
    \caption{We analyze the optimal number of parallel trajectories (base models) with respect to the epoch budget. Although additional trajectories improve diversity, they also incur extra warm-up cost. Empirically, a single base model is sufficient up to $\sim$120 epochs, while two base models become beneficial up to $\sim$240 epochs, with more trajectories favored at larger budgets. The dotted and dashed lines represent extrapolated data. The horizontal axis is the \emph{total} training budget $E$ across all $N$ base models, so each model is trained for $E/N$ epochs (e.g., $N{=}4$ at $E{=}32$ means four models trained for $8$ epochs each).
}
    \label{fig:compute-allocation}
\end{figure}

For a fixed compute budget $E$ (total training epochs),
hyper-epoch pretraining exposes three configuration knobs: epochs per
cycle $\tau$, cycles per base model $C$, and number of parallel base
models $N$, with $E\!=\!N\!\cdot\!C\!\cdot\!\tau$. We found in our experiments
that $\tau\!=\!2$ epochs/cycle consistently gave the best results.
We therefore fix $\tau\!=\!2$ for the main sweep and vary only the two
remaining knobs $(N, C)$. The X-axis of
Fig.~\ref{fig:compute-allocation} is therefore total training epochs
$E$, and each curve corresponds to a fixed $N$ traced out by varying
$C$, covering $E\!=\!32$ up to $E\!=\!960$. Under each budget tier we
report the configuration that minimises validation loss.

The optimal allocation tracks a clear staircase: small budgets prefer a
single base model with many cycles, and the optimal $N$ grows as $E$
grows. Empirically, the regime transitions occur at roughly geometric
intervals i.e., $N\!=\!1$ wins for $E\!\lesssim\!128$, $N\!=\!2$ for
$128\!\lesssim\!E\!\lesssim\!256$, $N\!=\!3$ for
$256\!\lesssim\!E\!\lesssim\!512$, and so on. As a rough empirical rule
of thumb (\textit{not} a precise law, the breakpoints in
Fig.~\ref{fig:compute-allocation} drift somewhat from a clean power of
two), the trend is well summarised by
\begin{equation}
    N^\star(E) \;\approx\; \max\!\Bigl(1,\; \bigl\lfloor \log_2\!\bigl(E / E_0\bigr) \bigr\rfloor\Bigr),
    \qquad E_0 \approx 64,
    \label{eq:n-star}
\end{equation}
i.e.\ each additional parallel base requires roughly \emph{doubling}
the epoch budget. We stress that Eq.~\ref{eq:n-star} should be read as
a qualitative guide rather than a tight fit; in particular, the
marginal benefit of going from $N$ to $N\!+\!1$ shrinks sharply past
$N\!\approx\!3$. Beyond three parallel bases the curves in
Fig.~\ref{fig:compute-allocation} bunch together and the ``winner'' at
each budget is decided by gaps on the order of $10^{-3}$ in validation
loss. Intuitively, every trajectory (base model) trained from random initialization pays
a fixed warm-up cost before its snapshots become useful members of the
ensemble; if the per-base budget $E/N$ is below this warm up cost,
additional parallel bases waste compute. Conversely, pushing a single
base to very long cycle counts is also wasteful: late cycles drift away
from good minima, the resulting snapshots are individually worse, and
the diversity inside the candidate pool degrades.
Eq.~\ref{eq:n-star} captures both ends of this trade-off; past
$N\!\approx\!3$, the trade-off is essentially saturated for our model
and data scale.

\begin{table}[!ht]
\centering
% \small
\caption{Snapshot placement comparison under small epoch ($E$) budgets. Snapshot methods use a single training trajectory, whereas independently trained models are specified explicitly.}
\label{tab:small-budget}
\begin{tabular}{clcc}
\toprule
$E$ & Configuration & Snapshots & Val.\ loss \\
\midrule
1 & normal training                     & 1 & 3.802 \\
1 & normal training $+$ 4 snapshots in last $0.15$ ep      & 4 & \textbf{3.765} \\
\midrule
4 &  normal training & 1 & 3.365 \\
4 & normal training (4 models trained independently for 1 ep each) & 4 & 3.687\\
4 & normal training (2 models trained independently for 2 ep each) & 2 & 3.501    \\
4 & snapshots at 2 ep/cycle    & 2 & 3.360 \\
4 & 3 ep normal training $+$ 4 snapshots in last 1 ep  & 4 & \textbf{3.326} \\
\bottomrule
\end{tabular}
\end{table}

\paragraph{Small epoch budget regime}
The geometric staircase of Eq.~\ref{eq:n-star} only kicks in once $E$
is large enough to support multiple base models and small cycles. In the practically
common low-budget regime $E\!\le\!4$, the question becomes \emph{how} to distribute snapshots
within that single trajectory. We find that concentrating many short
snapshots at the \emph{end} of training consistently outperforms the standard single checkpoint baseline.
Table~\ref{tab:small-budget} reports the best configurations we found
at each budget. At $E\!=\!1$, grabbing four closely-spaced
end-of-training snapshots performs better than a single checkpoint baseline. At $E\!=\!4$, dedicating the bulk of the
budget to a single trajectory and only diversifying at the end
(3 epochs normal $+$ 4 end-of-training snapshots in last epoch)
outperforms single checkpoint at the end, even 2 epochs/cycle (with 2 cycles) setting, and other baselines as shown in Table \ref{tab:small-budget}. However, when we move to an 8 epoch budget setting, our 2 ep/cycle setting with 4 cycles matches multiple snapshots at the end setting. In short, end-of-training snapshotting
  is the right primitive when the budget is too small for the cyclic
  schedule to amortise its overhead; the chain-distilled cyclic
  schedule takes over once $E$ is large enough to fit multiple
  genuine cycles.

%% ============================================================================
\section{Ablations and Analysis}
\label{sec:ablations}
The snapshot ensemble primitive is what delivers q0's speedups, which we
demonstrated in \S\ref{sec:compute-allocation}. We now analyze the
contributions of the other two components of our recipe, chain distillation and
the learned generalization prior.

\subsection{Contribution of Chain Distillation}
  \label{sec:abl-distillation}

\begin{table}[!ht]
\centering
\caption{Removing chain distillation from the full recipe (fixing
$\alpha = 0$ in Eq.~\ref{eq:chain-distill}); all other components
unchanged.}
\label{tab:distillation-ablation}
\vspace{0.5em}
\begin{tabular}{@{}lcc@{}}
\toprule
\textbf{Configuration}  & \textbf{Val Loss}  \\
\midrule
Full recipe                        & $\mathbf{3.0034}$ \\
\,$-$ w/o Chain distillation            & $3.0283$          \\
\bottomrule
\end{tabular}
\end{table}

We disable the chain distillation term in
Eq.~\ref{eq:chain-distill} (i.e.\ fix $\alpha = 0$ on every cycle) so
each snapshot is supervised only by the one-hot labels, and leave all
other components like parallel trajectories, the cyclic
snapshot collection schedule, epoch budget, and the learned prior selection at
inference unchanged. Table~\ref{tab:distillation-ablation} shows that removing chain distillation degrades the ensemble (at $256$ epochs), and is therefore
important at large epoch counts. Empirically, chain distillation draws consecutive
snapshots closer together and reduces within-trajectory disagreement. The
ensemble still improves, because the gain in individual quality outweighs the
lost diversity, as the error-ambiguity decomposition \citep{krogh1995neural}
predicts.

  % The
  % direction matches the claim from \S\ref{sec:method-chain-distill}:
  % the predecessor's output distribution carries dark knowledge
  % supervision that one-hot labels alone cannot transmit
  % \citep{furlanello2018born}. The price is a small drop in
  % within-trajectory diversity, but the gain in per snapshot quality
  % outweighs it, consistent with the error-ambiguity decomposition.

  %  The
  % direction matches the claim from \S\ref{sec:method-chain-distill}:
  % the predecessor's output distribution carries dark knowledge
  % supervision that one-hot labels alone cannot transmit
  % \citep{furlanello2018born}. 

\subsection{Anatomy of the Learned Prior}
  \label{sec:abl-selection}

\begin{figure}[!t]
\centering
\includegraphics[width=\textwidth]{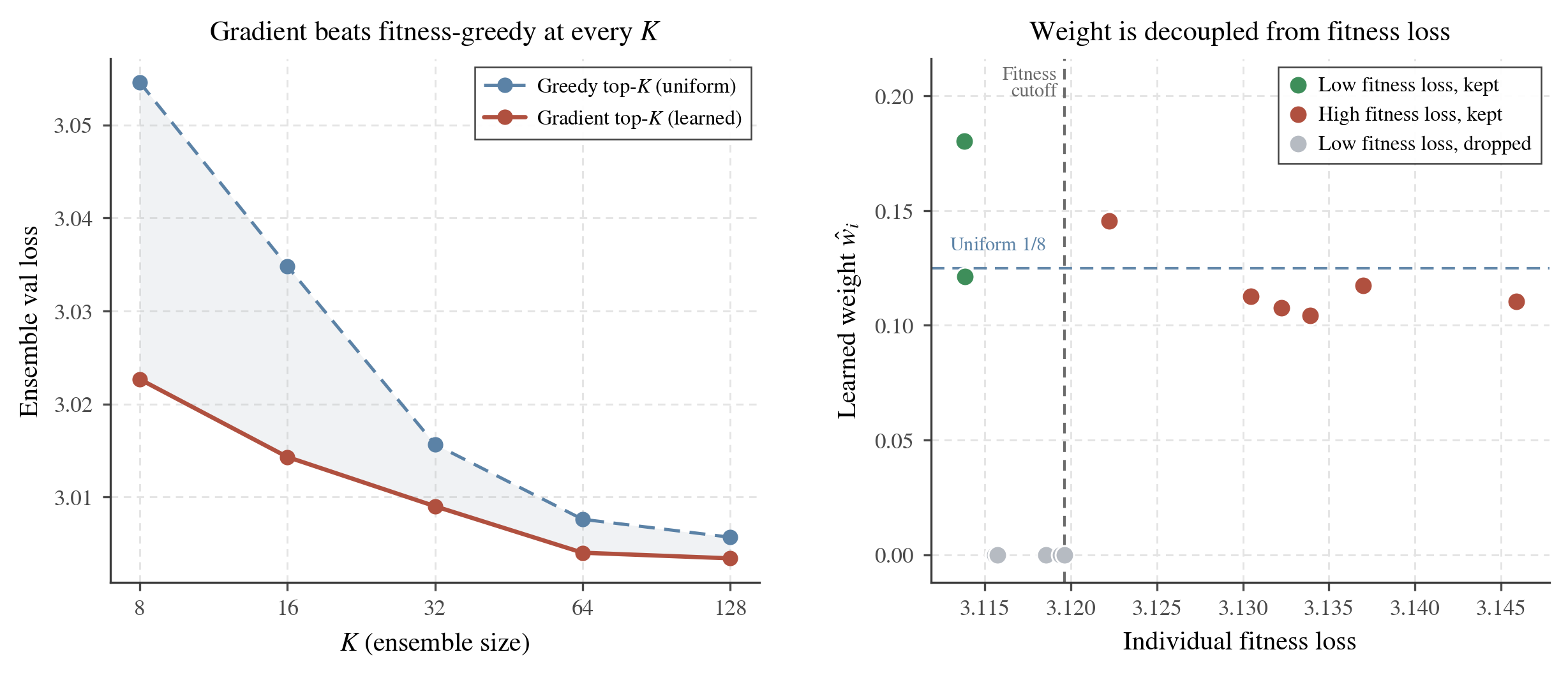}
\caption{Anatomy of the learned generalization prior for our
256 epoch run: \textbf{(left)} the prior beats a fitness-greedy
baseline at every $K$ with the largest margin at small $K$. At $K=8$, \textbf{(right)} fitness greedy picks
snapshots with lowest fitness set loss with uniform weights and
the prior selects snapshots
whose individual loss often exceeds the fitness greedy selections and non-trivial learned weights.}
\label{fig:selection-analysis}
\end{figure}

The learned generalization prior beats a \emph{fitness greedy} baseline
(which keeps the $K$ snapshots with the lowest individual fitness loss and
averages them uniformly) at \emph{every} inference budget $K$, and its
margin is largest at small $K$ (Figure~\ref{fig:selection-analysis}, left). At inference each ensemble member costs an additional forward pass, so per token cost grows linearly in $K$. The prior provides its largest gains at small inference budgets, which are the budgets most relevant in practice. The advantage comes from selecting for
complementary generalization rather than individual quality. At $K=8$ the two
methods share only $2$ of $8$ snapshots (Figure~\ref{fig:selection-analysis},
right). Fitness greedy piles onto the eight lowest loss checkpoints, whereas the prior keeps two of these and spends the rest on snapshots whose individual fitness loss is well above the
greedy cutoff but whose predictions cover errors the others miss. It also re-weights each member by its marginal contribution, the behavior induced by the non-separable mixture loss of Eq.~\ref{eq:learned-prior}.
  
% Does the prior pick the $K$ individually best snapshots, or does it
% select for complementary generalization? We compare the gradient learned generalization prior
% against a fitness greedy baseline that picks the $K$ snapshots with the
% lowest individual fitness loss and combines them uniformly. We find that our generalization prior performs better at every $K$, and that its advantage is largest at small $K$ (Figure~\ref{fig:selection-analysis}, left). At $K=8$ the two selections share only $2$ of $8$ picks. Fitness-greedy collapses on eight consecutive low fitness loss checkpoints,
% whereas the prior retains two of those and adds six other snapshots whose individual fitness loss is well above and is
% still selected (Figure \ref{fig:selection-analysis}, right). It thus performs two operations
% the fitness loss based baseline cannot i.e, (i) it includes snapshots whose
% individual quality is comparatively poor when their predictions
% cover errors the rest of the ensemble makes, and (ii) re-weights them in
% proportion to their marginal contribution, the behavior the
% non-separable mixture loss of Eq.~\ref{eq:learned-prior} induces.

\section{Related Work}
\label{sec:related}

\textbf{Scaling under data constraints.}
Neural scaling laws assume an effectively unbounded text supply
\citep{kaplan2020scaling, hoffmann2022training}, an assumption that breaks as
corpora approach the limits of available high quality data
\citep{villalobos2024position}. In the data constrained regime,
\citet{muennighoff2023scaling} find repeated epochs nearly as valuable as fresh
tokens up to roughly four passes, beyond which their value decays. Closest to ours, \citet{kim2025infinite} also study the compute-rich, data-limited regime and improve data efficiency by ensembling and heavy regularization of
models trained on the fixed corpus. We share the regime but spend the compute
differently by replacing independently trained members with cyclic trajectory snapshots, within trajectory
chain distillation, and a learned mixture prior.

\textbf{Ensembles of neural networks.}
An ensemble beats its members only when they are both accurate and diverse. The
error--ambiguity decomposition of \citet{krogh1995neural} formalizes this.
Deep ensembles obtain diversity from independent initializations that reach
distinct modes of the loss landscape \citep{lakshminarayanan2017simple,
fort2019deep}. Checkpoints from a single run cannot match this diversity because
they remain in one mode. Independent runs are also what pre-training cannot
afford. A fixed budget yields members of similar quality however many are
trained \citep{bouthillier2021accounting, summers2021nondeterminism}.

\textbf{Cheap ensembling within a trajectory.}
A complementary line of research extracts several models from one run. Warm restart and
cyclic learning rate schedules \citep{loshchilov2017sgdr} drive the optimizer
through multiple low loss basins, and snapshot ensembles
\citep{huang2017snapshot} ensemble the parameters saved at each cycle's end at
the cost of a single run. Fast Geometric Ensembling instead gathers members
along the low loss curves connecting minima
\citep{garipov2018loss, draxler2018essentially}. Weight space averaging such as 
Stochastic Weight Averaging \citep{izmailov2018averaging} and model soups
\citep{wortsman2022model} is cheaper but still requires a shared basin, which
our cross-trajectory members lack. \citet{smith2022general} cycle weight decay
\emph{in phase} with the learning rate; we instead anti-correlate the two, separating each cycle into exploratory and
exploitation phases.

\textbf{Distillation and self-training.}
Our chain distillation primitive draws on knowledge distillation
\citep{hinton2015distilling} and self-distillation, whose mechanism and link to
language model self-training we discuss in §\ref{sec:method-chain-distill}; here
we position it against the closest methods, which also distill \emph{within} a
single run. Snapshot Distillation \citep{yang2019snapshot} uses the last
snapshot of each cycle as the teacher for the next, and
\citet{wang2022efficient} find intermediate checkpoints to be effective
teachers. Both, however, use the in-trajectory teacher to regularize a single
final model and discard the intermediate snapshots. We invert this objective by keeping every snapshot as an ensemble member, so
that conditioning each cycle on its predecessor lifts the quality of every
member rather than only the final one. We pair this with cross-trajectory
diversity (§\ref{sec:method-snapshots}) and learned weighting
(§\ref{sec:method-weighting}).

\textbf{Selecting and weighting ensemble members.}
Most ensembles weight their members uniformly. This includes snapshot ensembles
and the deep ensembles of \citet{kim2025infinite}. Learned combination is
long established. Stacked generalization fits a combiner on held-out predictions
\citep{wolpert1992stacked}, and Bayesian model averaging weights predictors by
their posterior plausibility \citep{mackay1992bayesian, neal1996bayesian}. Our
learned prior (§\ref{sec:method-weighting}) is a lightweight instance of this
idea. It fits one softmax weighting over the members by minimizing held-out
negative log-likelihood, and reuses that single fit to select and weight members
at every inference budget.

%% ============================================================================

\section{Limitations}
\label{sec:limitations}

\textbf{Inference overhead.} Our method assumes that the inference cost of ensembling is acceptable. It
delivers its gains as an ensemble of $K$ snapshots, so inference costs $K$
forward passes rather than one. Smaller $K$ recovers most of the gain, but the
overhead is still prohibitive in many deployment settings. This is not
fundamental. The ensemble can be distilled into a single student that runs at
the inference cost of one model \citep{hinton2015distilling, kim2025infinite}.
This is compatible with our recipe, though we do not pursue it here.
  
\textbf{Training overhead.} Chain distillation adds a further
overhead, a teacher forward pass per cycle but without
any backward pass and is therefore highly optimizable.
Since the teacher is frozen within a cycle, its predictions on each example are
constant across that cycle's epochs and can be cached rather than recomputed.
Our preliminary experiments suggest caching across cycles may also be possible through a slower refresh of the teacher model, though we don't pursue that here and leave as a promising hypothesis. The learned prior adds only negligible overhead.
The fit operates on cached snapshot predictions rather than the models
themselves and is run once on a small fitness set.

\section{Conclusions}
We introduced hyper-epoch pretraining (\emph{q0}), a method that turns a
multi-epoch budget into a population of complementary models. Conceptually, the
biggest shift is away from a single refined model and toward superposing a
population of diverse models that fit the data. We believe
this matters in the regime where the data is fixed, compute is abundant, and
extra epochs no longer help a single model. Empirically, \emph{q0} matches a
strong ensemble baseline using far fewer epochs, and the gains hold across the full range of budgets we tested, from a
single epoch up to the largest. We also give prescriptive recipes for how to
allocate a given budget, since the best configuration shifts with scale. We hope \emph{q0} encourages further work on multi-epoch pretraining as a way of scaling compute over limited data, especially by efficiently searching for diverse models. Searching over a population is a different problem than training a single model with gradient descent, and likely calls for new primitives; the three we introduce here are a first set that give large gains, and we expect much more lies ahead.

\section{Acknowledgements}

We thank Andrew Gordon Wilson, Ethan Baron, and Hrithik Ravi for their valuable feedback on the
paper.

\section*{Author Contributions}
\label{sec:contributions}
\begin{itemize}[leftmargin=*, itemsep=2pt, topsep=2pt]
  \item \textbf{Bishwas Mandal}: developed chain distillation and snapshot
        ensemble; ran most experiments and co-wrote the paper.
  \item \textbf{Shmuel Berman}: developed snapshot ensemble; provided feedback
        on the paper.
  \item \textbf{Akshay Vegesna}: developed the learned generalization prior;
        built small-scale baselines for fast iteration; ran downstream
        benchmarking and helped with experimentation.
  \item \textbf{Samip Dahal}: led the project; directed experiments; shaped the
        core framing of the paper and helped write it.
\end{itemize}

% \section*{Author Contributions}
% \label{sec:contributions}

% \textbf{Bishwas Mandal}: developed chain distillation and snapshot
% ensemble; ran most experiments and co-wrote the paper.

% \textbf{Shmuel Berman}: developed snapshot ensemble; provided feedback
% on the paper.

% \textbf{Akshay Vegesna}: developed the learned generalization prior; built small-scale baselines for fast
% iteration; ran downstream benchmarking and helped with experimentation.

% \textbf{Samip Dahal}: led the project; directed experiments; shaped
% core framing of the paper and helped write it.

\newpage
\bibliographystyle{unsrtnat}
\bibliography{references}

\newpage
\appendix

\section{Overview of hyper-epoch pretraining}
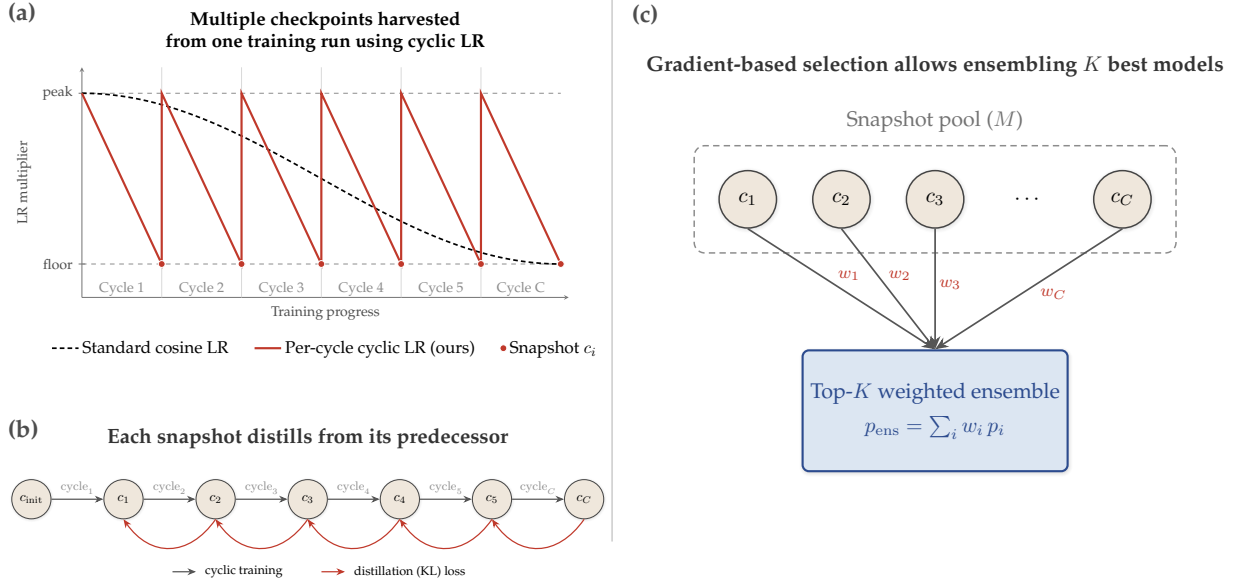
\begin{figure*}[!h]
    \centering
%% ============================================================
%% (a) LR schedule
%% ============================================================
\sbox\panelLR{%
\begin{tikzpicture}
\begin{axis}[
    width=11cm, height=5.2cm, scale only axis,
    xmin=0, xmax=6.1, ymin=-0.10, ymax=1.14,
    axis lines=left, axis line style={black!55},
    clip=false,
    xtick=\empty, ytick={0.08,1.0}, yticklabels={floor, peak},
    ytick style={black!55},
    yticklabel style={font=\footnotesize, text=black!70},
    xlabel={\footnotesize Training progress}, ylabel={\footnotesize LR multiplier},
    xlabel style={text=black!70}, ylabel style={text=black!70},
    title style={align=center, font=\fontsize{12}{14}\selectfont\bfseries},
    title={Multiple checkpoints harvested\\ from one training run using cyclic LR},
    legend style={at={(0.5,-0.16)}, anchor=north, legend columns=3, draw=none, fill=none,
                  font=\normalsize, /tikz/every even column/.append style={column sep=14pt}},
    legend cell align=left,
]
\draw[black!45, dashed, line width=0.6pt] (axis cs:0,1.0)  -- (axis cs:6,1.0);
\draw[black!45, dashed, line width=0.6pt] (axis cs:0,0.08) -- (axis cs:6,0.08);
\draw[black!15, line width=0.6pt] (axis cs:1,-0.10) -- (axis cs:1,1.14);
\draw[black!15, line width=0.6pt] (axis cs:2,-0.10) -- (axis cs:2,1.14);
\draw[black!15, line width=0.6pt] (axis cs:3,-0.10) -- (axis cs:3,1.14);
\draw[black!15, line width=0.6pt] (axis cs:4,-0.10) -- (axis cs:4,1.14);
\draw[black!15, line width=0.6pt] (axis cs:5,-0.10) -- (axis cs:5,1.14);
\node[font=\footnotesize, text=black!45, anchor=north] at (axis cs:0.5,0.015) {Cycle 1};
\node[font=\footnotesize, text=black!45, anchor=north] at (axis cs:1.5,0.015) {Cycle 2};
\node[font=\footnotesize, text=black!45, anchor=north] at (axis cs:2.5,0.015) {Cycle 3};
\node[font=\footnotesize, text=black!45, anchor=north] at (axis cs:3.5,0.015) {Cycle 4};
\node[font=\footnotesize, text=black!45, anchor=north] at (axis cs:4.5,0.015) {Cycle 5};
\node[font=\footnotesize, text=black!45, anchor=north] at (axis cs:5.5,0.015) {Cycle C};
\addplot[black, densely dashed, line width=1.1pt, samples=200, domain=0:6]
    {0.08+0.92*0.5*(1+cos(deg(pi*x/6)))};
\addlegendentry{Standard cosine LR}
\addplot[distill, line width=1.4pt] coordinates {
   (0,1)(1,0.08)(1,1)(2,0.08)(2,1)(3,0.08)(3,1)(4,0.08)(4,1)(5,0.08)(5,1)(6,0.08)};
\addlegendentry{Per-cycle cyclic LR (ours)}
\addplot[only marks, mark=*, mark size=2.2pt, mark options={fill=distill, draw=white, line width=0.5pt}]
    coordinates {(1,0.08)(2,0.08)(3,0.08)(4,0.08)(5,0.08)(6,0.08)};
\addlegendentry{Snapshot $c_i$}
\end{axis}
\end{tikzpicture}%
}
%% ============================================================
%% (b) Chain distillation
%% ============================================================
\sbox\panelChain{%
\begin{tikzpicture}[
  >={Stealth[length=2.2mm,width=1.8mm]},
  snap/.style={circle, draw=nodeedge, fill=nodefill, line width=0.7pt,
               minimum size=10mm, font=\small,
               blur shadow={shadow blur steps=4, shadow xshift=0.3pt, shadow yshift=-0.5pt, shadow opacity=18}},
  fwd/.style={->, draw=arrowcol, line width=0.9pt},
  dist/.style={->, draw=distill, line width=0.9pt},
  clab/.style={above, font=\footnotesize, text=black!45},
  ]
  \def\dx{2.35}
  \node[snap] (n0) at (0,0)        {$c_{\mathrm{init}}$};
  \node[snap] (n1) at (\dx,0)      {$c_1$};
  \node[snap] (n2) at (2*\dx,0)    {$c_2$};
  \node[snap] (n3) at (3*\dx,0)    {$c_3$};
  \node[snap] (n4) at (4*\dx,0)    {$c_4$};
  \node[snap] (n5) at (5*\dx,0)    {$c_5$};
  \node[snap] (n6) at (6*\dx,0)    {$c_C$};
  \node[font=\fontsize{15}{17}\selectfont\bfseries, text=black!80] at (3*\dx,1.55)
        {Each snapshot distills from its predecessor};
  \draw[fwd] (n0) -- node[clab]{$\mathrm{cycle}_1$} (n1);
  \draw[fwd] (n1) -- node[clab]{$\mathrm{cycle}_2$} (n2);
  \draw[fwd] (n2) -- node[clab]{$\mathrm{cycle}_3$} (n3);
  \draw[fwd] (n3) -- node[clab]{$\mathrm{cycle}_4$} (n4);
  \draw[fwd] (n4) -- node[clab]{$\mathrm{cycle}_5$} (n5);
  \draw[fwd] (n5) -- node[clab]{$\mathrm{cycle}_C$} (n6);
  \foreach \a/\b in {n2/n1, n3/n2, n4/n3, n5/n4, n6/n5}{
     \draw[dist] (\a.south) to[out=-125,in=-55,looseness=1.3] (\b.south);
  }
  \def\ly{-1.85}
  \draw[fwd] (3.6,\ly) -- (4.2,\ly);
  \node[right=1mm, font=\footnotesize, anchor=west] at (4.2,\ly) {cyclic training};
  \draw[dist] (7.4,\ly) -- (8.0,\ly);
  \node[right=1mm, font=\footnotesize, anchor=west] at (8.0,\ly) {distillation (KL) loss};
\end{tikzpicture}%
}
%% ============================================================
%% (c) Learned selection / ensemble
%% ============================================================
\sbox\panelSel{%
\begin{tikzpicture}[
  >={Stealth[length=2.2mm,width=1.8mm]},
  snap/.style={circle, draw=nodeedge, fill=nodefill, line width=0.7pt,
               minimum size=9.5mm, font=\small,
               blur shadow={shadow blur steps=4, shadow xshift=0.3pt, shadow yshift=-0.5pt, shadow opacity=18}},
  ebox/.style={rounded corners=3pt, draw=ensedge, fill=ensfill, line width=1pt,
               align=center, font=\small, text=ensedge, minimum width=40mm, minimum height=20mm,
               blur shadow={shadow blur steps=5, shadow xshift=0.4pt, shadow yshift=-0.7pt, shadow opacity=22}},
  flow/.style={->, draw=arrowcol, line width=0.8pt},
  ]
  \def\dx{1.55}
  \def\py{1.0}
  \node[snap] (c1) at (0,\py)        {$c_1$};
  \node[snap] (c2) at (\dx,\py)      {$c_2$};
  \node[snap] (c3) at (2*\dx,\py)    {$c_3$};
  \node[font=\small] at (3*\dx,\py)  {$\cdots$};
  \node[snap] (cC) at (4*\dx,\py)    {$c_C$};
  \node[rounded corners=5pt, draw=black!45, densely dashed, line width=0.7pt,
        fill=none, fit=(c1)(cC), inner xsep=4mm, inner ysep=4mm] (pool) {};
  \node[above=0.8mm, font=\small, text=black!55] (plab) at (pool.north) {Snapshot pool ($M$)};
  \node[above=3mm, font=\fontsize{10}{12}\selectfont\bfseries, text=black!80] at (plab.north)
        {Gradient-based selection allows ensembling $K$ best models};
  \node[ebox] (ens) at (2*\dx,-2.5) {Top-$K$ weighted ensemble\\[4pt]
        $p_{\mathrm{ens}} = \sum_{i} w_i\, p_i$};
  \foreach \n/\w in {c1/w_1,c2/w_2,c3/w_3,cC/w_C}{
     \draw[flow] (\n.south) -- node[pos=0.46, auto, font=\scriptsize, text=wcol,
                 inner sep=1.6pt]{$\w$} (ens.north);
  }
\end{tikzpicture}%
}
%% ============================================================
%%  Assembly:  (a) top-left, (b) bottom-left, (c) right
%% ============================================================
\resizebox{\textwidth}{!}{%
\begin{tikzpicture}
  \node[anchor=north west] (lr) at (0,0)      {\resizebox{8.6cm}{!}{\usebox\panelLR}};
  \node[anchor=north west] (ch) at (0,-6.05)  {\resizebox{8.6cm}{!}{\usebox\panelChain}};
  \node[anchor=north west] (se) at (9.1,-0.6) {\resizebox{8.6cm}{!}{\usebox\panelSel}};
  \draw[black!18, line width=0.7pt] (8.85,0.1) -- (8.85,-7.9);
  \node[font=\small\bfseries, anchor=north west, text=black!70] at (-0.05,0.05)  {(a)};
  \node[font=\small\bfseries, anchor=north west, text=black!70] at (-0.05,-6.00) {(b)};
  \node[font=\small\bfseries, anchor=north west, text=black!70] at (9.0,0.02)   {(c)};
\end{tikzpicture}
}
    \caption{(a) our cyclic learning rate schedule: a single run is split into $C$ cycles, each restart-and-anneal yielding one snapshot $c_i$, versus a standard single-cycle cosine. (b) chain distillation within a trajectory: each snapshot trains with an additional KL term against its frozen predecessor. (c) a learned prior weights the snapshot pool, and the top-$K$ are combined as $p_{\mathrm{ens}}=\sum_i w_i\,p_i$.}
    \label{fig:schedule}
 \end{figure*}

\section{Implementation Details}
\label{app:hyperparams}

\paragraph{Reported configurations}
We report two configurations that share the same per-step recipe and
differ only in how the training budget is partitioned into
trajectories $N$ and cycles per trajectory $C$:

\begin{itemize}
\item \textbf{q0 (256 epoch headline)} $N = 2$ parallel trajectories
    of $C = 64$ cycles at $2$ epochs per cycle, giving
    $M = N \cdot C = 128$ snapshots and $128$ epochs of training
    per trajectory.
\item \textbf{q0 scaled.} $N = 5$ trajectories of $C = 96$ cycles,
    giving $M = 480$ snapshots. All other hyperparameters are
    identical.
\end{itemize}

Trajectories run in parallel, one per compute node ($8$ GPUs each
under intra-node DDP); there is no inter-trajectory gradient
synchronisation.

\paragraph{Model architecture}
$1.8$B parameter decoder-only transformer with $n_{\mathrm{layer}} = 30$,
$n_{\mathrm{embd}} = 2048$, $n_{\mathrm{head}} = 16$, sequence length
$2048$, SwiGLU MLP, U-Net skip connections from encoder to decoder
layers, attention gating, and value-embedding projections.

\paragraph{Layer duplication (looping)}
The decoder layers with indices in $[15, 25)$ (i.e., the 10 layers
$15, 16, \newline \dots, 24$) form a \emph{looped middle block}: from the very
first training step (in q0 setting), the forward pass traverses this block
\emph{five} times before continuing to the final five decoder layers
$25, \dots, 29$. Concretely, the decoder half is computed as
\[
  \underbrace{15 \rightarrow 24}_{\text{pass 1}}
  \;\rightarrow\;
  \underbrace{15 \rightarrow 24}_{\text{pass 2}}
  \;\rightarrow\;\cdots\;\rightarrow\;
  \underbrace{15 \rightarrow 24}_{\text{pass 5}}
  \;\rightarrow\;
  25 \rightarrow 29 ,
\]
giving the model an extra $4\times$ depth through that middle block
at no parameter cost. U-Net skip connections from the corresponding
encoder layers, the residual-/$x_0$-mixing scalars
$(\lambda^{\mathrm{res}}_i, \lambda^{x_0}_i)$, and the
value-embedding projections are re-applied on every pass through
the looped layers. The looping configuration (start/end indices,
number of repetitions) is fixed for the entirety of training and is
identical across all snapshots and across both reported
configurations.  For the baseline, looping is applied to each model over exactly the same
layers described above, but it is enabled only after the first $50\%$ of that
model's training rather than from the first step. Applying looping for the
entire training run does not further improve the baseline.

\paragraph{Optimizer}
Training uses a Muon~$+$~AdamW hybrid: Muon for the $2$D matrix
parameters, AdamW for embeddings, the LM head, and the scalar/skip
parameters. Peak learning rates are $0.04$ (matrix), $0.1$ (scalar),
$0.15$ (embedding), $0.002$ (unembedding), each scaled by a global
multiplier of $0.25$; peak weight decay is $1.3$. AdamW uses
$(\beta_1, \beta_2) = (0.8, 0.95)$; Muon uses momentum $0.95$ with
$5$ Newton-Schulz iterations. Gradients are clipped to global norm
$1.0$ before every optimizer step.

\paragraph{Multi-token prediction auxiliary loss}
The final hidden state is also routed through a single extra
transformer block, projected via a learned linear map of the
next-token embedding concatenated with the residual, and supervised
to predict the token at position $t+1$. The auxiliary cross-entropy
loss is weighted by $\lambda_{\mathrm{MTP}} = 0.3$ and added to the
primary language-modelling loss. When chain distillation is active
(see below) the MTP term is folded into the ``normal loss'' side of
the distillation trade-off so that the effective objective is
\[
    \mathcal{L} \;=\;
    (1 - \alpha)\,\bigl(\mathcal{L}_{\mathrm{CE}} +
                        \lambda_{\mathrm{MTP}} \mathcal{L}_{\mathrm{MTP}}\bigr)
    \,+\,
    \alpha\, T^2 \,\mathcal{L}_{\mathrm{KL}}.
\]

\paragraph{Cyclic LR/WD schedule}
For a cycle of length $L$ steps, the per-step LR and WD multipliers
applied on top of the peak optimizer values are
\begin{equation}
\label{eq:cyclic-schedule}
    m_{\mathrm{LR}}(s)
    \,=\, f_{\mathrm{LR}} + (1 - f_{\mathrm{LR}})\!\cdot\!
        \bigl(1 - \tfrac{s}{L}\bigr),
    \qquad
    m_{\mathrm{WD}}(s)
    \,=\, f_{\mathrm{WD}} + (1 - f_{\mathrm{WD}})\!\cdot\!
        \frac{1 - m_{\mathrm{LR}}(s)}{1 - f_{\mathrm{LR}}},
\end{equation}
with $f_{\mathrm{LR}} = 0.08$ and $f_{\mathrm{WD}} = 0.7$. The LR
therefore sweeps $[0.08,\,1] \times$ peak linearly within each
cycle, and weight decay sweeps $[0.7,\,1] \times$ peak in the
opposite direction (an ``inverse-LR'' WD schedule).

\paragraph{Batch-size schedule}
Across cycles the batch size follows a block-triangular schedule:
the peak ($2^{19} = 524{,}288$ tokens) is used for the middle
${\sim}50\%$ of cycles, and a floor of $393{,}216$ tokens is used
for the early and late cycles. Batch size is held constant within a
cycle; only LR and WD vary intra-cycle.

\paragraph{Cycle-boundary perturbation}
At every cycle boundary we perturb each parameter as
\begin{equation}
\label{eq:perturb}
    \theta \;\leftarrow\;
    \theta \,+\, \sigma_c \cdot \mathrm{std}(\theta) \cdot z,
    \qquad z \sim \mathcal{N}(0,\,I),
\end{equation}
so cycle $c$'s student starts from a perturbed copy of
$\theta_{n,c-1}$ rather than from $\theta_{n,c-1}$ itself, recovering
a small amount of within-trajectory diversity that chain distillation
would otherwise remove. The magnitude $\sigma_c$ is cosine-decayed
from $\sigma_{\max} = 0.25$ at the first cycle boundary to
$\sigma_{\min} = 0.05$ at the last. Noise is sampled independently
per parameter, scaled to that parameter's own running standard
deviation, and is shared across all GPUs in a node (deterministic
per-cycle seed) so DDP replicas remain bit-identical after the
perturbation.

\paragraph{Chain distillation}
We use $\alpha = 0.45$, $T = 1.2$, and $c_{\mathrm{start}} = 8$
warmup cycles before distillation engages. The teacher is always the
most recently saved snapshot $\theta_{n,c-1}$ (refreshed at the end
of every cycle); the KL term is computed in the
$\mathrm{KL}\!\bigl(\sigma(f_{\mathrm{teacher}}/T)\,\|\,
\sigma(f_{\mathrm{student}}/T)\bigr)$ direction. The teacher is
frozen, kept in bf16 with the same layer-duplication configuration
as the student, and runs under \texttt{torch.no\_grad}. We swept
$\alpha \in \{0.20, 0.30, 0.40, 0.45, 0.50, 0.55, 0.60, 0.65, 0.70\}$
and $T \in \{0.8, 0.9, 1.0, 1.1, 1.2, 1.3, 1.4\}$ on the
$256$-epoch budget; the reported $(0.45,\,1.2)$ pair was the
optimal configuration.

\paragraph{Data and fitness set}
Training data is FineWeb. From the beginning of the training file we
carve out the first ${\approx}132\text{K}$ tokens at whole-document
granularity (no document is split) and remove that prefix from the
training stream of \emph{every} trajectory; this prefix is the
fitness set $\mathcal{F}$ used to fit the learned prior. The
remaining $100$M training tokens are shuffled (same shuffle across
trajectories, different per-cycle stream ordering) and consumed at
two epochs per cycle. Validation $\mathcal{V}$ is a disjoint $10$M token
set from the same distribution.

\paragraph{Learned prior fit}
The mixture logits $\boldsymbol{\beta} \in \mathbb{R}^M$ are
optimised on a single node by minimising
Eq.~\ref{eq:learned-prior} on the cached $P_{\mathcal{F}}$ with
AdamW (learning rate $0.5$, weight decay $0$, gradient clipped at
norm $1.0$) for $300$ full-batch steps starting from
$\boldsymbol{\beta} \leftarrow 10^{-4}\,\mathcal{N}(0,\,I_M)$. The
final logits are broadcast to all nodes so every node has
bit-identical selections.

\paragraph{Ensemble evaluation}
Given the learned $w^\star = \mathrm{softmax}(\boldsymbol{\beta})$,
for each inference budget $K \in \{1, 2, 4, 8, 16, 32, 64, 128, 256\}$
(clamped to $M$) we take $\mathcal{S}_K$ as the top-$K$ indices of
$w^\star$, renormalise the weights inside $\mathcal{S}_K$ so they sum
to one, and evaluate
\[
    p_{\mathrm{ens}}(y_t \mid x_{<t})
    \;=\; \sum_{i \in \mathcal{S}_K} \tilde{w}_i \, p_i(y_t \mid x_{<t})
\]
on $\mathcal{V}$ by streaming each selected snapshot in turn and
accumulating the per-token mixture probability across nodes. The
ensemble validation loss is the negative log of this mixture
averaged over tokens.

\paragraph{Distributed setup}
Each trajectory occupies one $8$-GPU node. Within a node, gradient
synchronisation uses PyTorch DDP on an intra-node process group;
across nodes the only communication is (i) all-gathering the
$P_{\mathcal{F}}$ rows before the learned-prior fit, (ii)
broadcasting the fitted $\boldsymbol{\beta}$ back to every node, and
(iii) sharded ensemble evaluation in which the $K$ selected
snapshots are partitioned across nodes and per-token mixture
probabilities are all-reduced. Training is otherwise fully parallel across trajectories.

\paragraph{Tokenization and document packing}
We tokenize FineWeb with the GPT-2 BPE tokenizer (tiktoken
encoding gpt2, vocabulary size $50{,}257$). Each document is
prefixed with the \texttt{<|endoftext|>} token and concatenated into
one long stream; we record document boundaries during preprocessing
but do not apply doc-boundary attention masking at training time, so
the model sees standard packed sequences of length $2048$ with
\texttt{<|endoftext|>} acting as the only delimiter between
documents. The training stream is shuffled at the document level
with a fixed seed shared across trajectories so that every $\theta_i$
sees the same underlying token order modulo its cyclic stream
re-permutation.

\paragraph{Baseline}
The matched-compute baseline trains $N_b = 8$ independent models from
random initialization. Each baseline run uses the same architecture,
optimizer, peak learning rates, weight decay, batch size, MTP
auxiliary loss, and layer-duplication block as q0; the differences
are entirely in the training schedule. Each baseline model is
trained for $32$ epochs on the $100$M-token stream (so the eight
models together consume $256$ epochs, matching the $E = 256$ q0
configuration). The learning rate follows a single cosine decay
across all $32$ epochs of each model rather than a per-cycle cyclic
schedule; weight decay is held constant at its peak; batch size is
held at $524{,}288$ tokens throughout; and there is no chain
distillation, no cycle-boundary perturbation, and no learned prior.
Layer duplication on the baseline activates only for the final
$50\%$ of training rather than from step $0$, and dropout of $0.1$
is applied throughout to compensate for the longer effective
per-model exposure to the data. We additionally maintain an
exponential moving average of the parameters with decay $0.95$
during the last $10\%$ of each model's training run, and at
inference uniformly average the eight EMA-weight models in
probability space (softmax averaging, $w_i = 1/8$).

\paragraph{Downstream evaluation}
Zero-shot accuracy on ARC-Easy, PIQA, and SciQ is reported via the
EleutherAI \texttt{lm-evaluation-harness} using the default
\texttt{acc} metric (no length normalisation) and the harness's
standard prompts for each task. We evaluate the same top-$K$
weighted snapshot ensemble that minimises validation loss on
$\mathcal{V}$ (Table~\ref{tab:main-result}); per-task scores are
obtained by summing the per-class log-probabilities across the
selected snapshots after weighting by $\tilde w_i$ and then taking
the argmax.

\paragraph{Number of seeds}
Apart from the downstream benchmark accuracies in
Table~\ref{tab:benchmarks}, which report mean and standard error
across multiple harness evaluations of the same trained ensemble,
all numerical results in the paper come from a single training run
per configuration. We confirmed in spot checks during development
that re-runs at fixed hyperparameters produced ensemble validation
losses within ${\sim}10^{-3}$ of one another, but we do not report
seed variance in the main tables.

\paragraph{Reproducibility}
The full training, snapshot caching, learned-prior fit, and
ensemble evaluation code is released here: \url{https://github.com/qlabs-eng/slowrun/commit/a200723a72510f51adcec85224933d4c205a53fb}.

\section{Slowrun Data-Efficiency Analysis}
\label{app:slowrun}

The Slowrun challenge benchmarks language modeling algorithms in the
fixed-data, infinite-compute regime ($100$M FineWeb tokens, no
compute or time limit, lowest validation loss wins), with the
explicit goal of admitting better algorithm classes that wall clock
speedrun benchmarks filter out. Our setup matches its data budget.
For a given validation loss, the setting uses piecewise-linear
interpolation along the nanochat reference curve in
Table~\ref{tab:slowrun-reference} (built by taking the best
validation loss across nanochat sizes d12, d20, d26 at each token
count) to find the equivalent dataset size; data efficiency is the
ratio of that size to our $100$M-token budget.

\begin{table}[H]
\centering
\caption{Nanochat reference validation losses used in the Slowrun
data-efficiency setting (best across d12, d20, d26 at each token count).}
\label{tab:slowrun-reference}
\vspace{0.5em}
\begin{tabular}{@{}lc@{}}
\toprule
\textbf{Tokens} & \textbf{Best Val Loss} \\
\midrule
100M  & $4.7324$ \\
200M  & $3.8834$ \\
400M  & $3.3673$ \\
600M  & $3.1999$ \\
800M  & $3.1257$ \\
1.0B  & $3.0462$ \\
1.2B  & $3.0026$ \\
1.4B  & $2.9668$ \\
1.6B  & $2.9470$ \\
1.8B  & $2.9126$ \\
2.0B  & $2.8917$ \\
\bottomrule
\end{tabular}
\end{table}

Applying this to our $256$-epoch numbers: the baseline ($3.0476$)
interpolates between the $800$M and $1$B reference points to
$996.5$M equivalent tokens (${\approx}9.96\times$); q0 ($3.0034$)
interpolates between $1$B and $1.2$B to $1196.3$M
(${\approx}11.96\times$); and q0 scaled ($2.987$) interpolates
between $1.2$B and $1.4$B to $1287.2$M (${\approx}12.87\times$).
Our baseline is therefore already a strong, data-efficient reference
(matching the nanochat curve at $1$B tokens on a $100$M budget), so
the q0 and scaled-q0 gains are measured on top of a strong baseline
rather than a weak one; both are meaningful shifts given that the
reference curve is already flat in this regime (only ${\sim}0.08$
loss reduction across the range from $1$B to $1.4$B tokens).

\subsection{Downstream Data Efficiency}
\label{app:slowrun-downstream}
We apply the same interpolation to downstream accuracy. The reference curve in
Table~\ref{tab:slowrun-downstream-ref} gives the mean accuracy across runs at
each token count, used in place of the loss curve. For a given accuracy we find
the equivalent token count by piecewise-linear interpolation between the two
bracketing reference points and divide by our $100$M budget; this yields the
per-benchmark and average data-efficiency multipliers reported in
Table~\ref{tab:benchmarks}. On average accuracy the baseline reaches
${\sim}12.2\times$, q0 reaches ${\sim}14.2\times$, and the scaled variant
reaches ${\sim}16.0\times$, consistent with the loss-based ranking.

\begin{table}[H]
\centering
\caption{Nanochat reference downstream accuracy (mean across runs) used for the
downstream data-efficiency interpolation.}
\label{tab:slowrun-downstream-ref}
\vspace{0.5em}
\begin{tabular}{@{}lcccc@{}}
\toprule
\textbf{Tokens} & \textbf{PIQA} & \textbf{ARC-Easy} & \textbf{SciQ} & \textbf{Avg} \\
\midrule
100M  & 0.5586 & 0.3030 & 0.2977 & 0.3864 \\
200M  & 0.6086 & 0.3676 & 0.5740 & 0.5167 \\
400M  & 0.6426 & 0.4286 & 0.7070 & 0.5927 \\
600M  & 0.6533 & 0.4550 & 0.7520 & 0.6201 \\
800M  & 0.6701 & 0.4790 & 0.7577 & 0.6356 \\
1.0B  & 0.6723 & 0.4832 & 0.7740 & 0.6432 \\
1.2B  & 0.6763 & 0.4777 & 0.7747 & 0.6429 \\
1.4B  & 0.6767 & 0.5046 & 0.7897 & 0.6570 \\
1.6B  & 0.6788 & 0.5041 & 0.7967 & 0.6599 \\
1.8B  & 0.6859 & 0.5123 & 0.7993 & 0.6659 \\
2.0B  & 0.6919 & 0.5260 & 0.8087 & 0.6755 \\
\bottomrule
\end{tabular}
\end{table}

\end{document}